# A Hybrid Algorithm to Compute Marginal and Joint Beliefs in Bayesian Networks and Its Complexity

**Mark Bloemeke**
Artificial Intelligence Laboratory
Computer Science Department
University of South Carolina
bloemeke@cs.sc.edu

**Marco Valtorta**
Artificial Intelligence Laboratory
Computer Science Department
University of South Carolina
mgv@cs.sc.edu

## Abstract

There exist two general forms of exact algorithms for updating probabilities in Bayesian Networks. The first approach involves using a structure, usually a clique tree, and performing local message based calculation to extract the belief in each variable. The second general class of algorithm involves the use of non-serial dynamic programming techniques to extract the belief in some desired group of variables. In this paper we present a hybrid algorithm based on the latter approach yet possessing the ability to retrieve the belief in all single variables. The technique is advantageous in that it saves a NP-hard computation step over using one algorithm of each type. Furthermore, this technique re-enforces a conjecture of Jensen and Jensen [JJ94] in that it still requires a single NP-hard step to set up the structure on which inference is performed, as we show by confirming Li and D'Ambrosio's [LD94] conjectured NP-hardness of OFP.

## 1 Overview

Bayesian Networks(BN) provide a standard way to represent a probability distribution on a series of discrete propositional variables. By taking advantage of independence information between the variables, BN's can reduce the amount of space necessary to specify the distribution, but they then require special algorithms to recover meaningful distributions. One such algorithm to recover the marginals of all the variables is known as the tree of cliques approach [LS88] [Pea88] [Nea90] [Jen96].

Another approach to the calculation of a marginal probability distribution on a set of target variables, called Symbolic Probabilistic Inference (SPI) is discussed in [LD94]. It involves solving the Optimal Factoring Problem (OFP defined in Section 4) for the target set of variables whose distribution you are interested in. The solution to the OFP is then used to combine the conditional probability tables that describe the Bayesian Network and extract the desired marginal distribution.

Unknown, however, was the time complexity of the OFP. In [LD94] it was suggested that the OFP was NP-hard, but this was never shown. In sections 4 to 6 of this paper we will confirm Li and D'Ambrosio's conjecture that the OFP is indeed NP-hard by reduction from the secondary problem of non-serial dynamic programming.

In section 7 through 10 a new method, based on Li and D'Ambrosio's, is given that uses an OFP solution to build a data structure (called a factor tree, which is similar to the expression tree of [LD93]) from which not only the target joint belief can be extracted, but also all the marginal beliefs. This is obtained by using a method that is similar in outline to the tree of cliques approach. This similarity extends even to the complexity of the algorithm in such a way as to further confirm Jensen's hypothesis that all algorithms as efficient as the tree of cliques that recover single marginals must include an NP-hard step.

## 2 Symbolic Probabilistic Inference

Assuming that we have a Bayesian Network with DAG $G = (V, E)$ and conditional probability tables $P(v_i|\Pi(v_i))$, where $\Pi(v_i)$ are the parents of $v_i$ in $G$, we can, if only very inefficiently, recover the total joint probability using the chain rule for Bayesian Networks:

$$P(V) = \prod_{v_i \in V} P(v_i|\Pi(v_i)) \qquad (1)$$

and from this we can use marginalization to retrieve our belief in any subset of variables $V'$ as:

$$P(V') = \sum_{V-V'} P(V). \qquad (2)$$

The SPI algorithm is based on direct use of equations 1 and 2 to retrieve any desired joint. In order to avoid the expo-nential size of the resulting tables the fact that multiplication distributes over addition is employed to push some of the summations down into the products. This allows some control to be maintained over the size and time complexity of the resulting calculation by allowing variable elimination from the joint at the earliest possible time. The true cost of this method in fact hinges upon which ordering of terms is selected for equation 1.

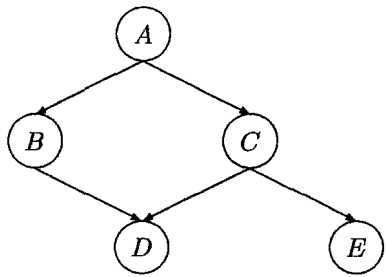

Figure 1: Simple Example Network.

For example consider the network shown in Figure 1. We can calculate the joint probability of the variables $A$ and $C$ directly from equations 1 and 2 using the equation

$$P(A,C) = \sum_{B,D,E} P(E|C) * P(D|B,C) * P(C|A) * P(B|A) * P(A). \quad (3)$$

Assuming that each variable $A, B, C, D, E$ has two states, this will need a table with $2^5$ entries to be calculated that will requires at least $2^2 + 2^3 + 2^4 + 2^5$ multiplications to construct and 28 additions to marginalize onto $A$ and $C$. Thus using just equations 1 and 2 to get $P(A,C)$ will require a total of 92 significant operations.

However, with a slight re-ordering of the terms combined by equation 1 followed by the distribution of the summations from 2, we get

$$P(A,C) = P(A) * [P(C|A) * \sum_E [P(E|C) * [\sum_B P(B|A) * \sum_D P(D|B,C)]]] \quad (4)$$

which requires only 24 multiplications and 12 additions for a total of 36 significant operations.

Since we can only push the summation of a variable down as far as its earliest occurrence in the combination ordering, the ordering determines the amount of time and space we can save. An appropriate combinatorial optimization approach is defined in [LD94] that treats each conditional probability table as a set of variables and defines a combination function for two sets and a cost function based on combination. Then the optimal set combination ordering with respect to cost function minimality can be derived for any set of target variables whose joint density is required. From that ordering the calculation of the joint occurs in accordance with equations 1 and 2 utilizing the distribution described above.

| $A$ | $B$ | $f_2(A,B)$ |
|---|---|---|
| $a$ | $b$ | 3 |
| $a$ | $\neg b$ | 4 |
| $\neg a$ | $b$ | 1 |
| $\neg a$ | $\neg b$ | 5 |

| $B$ | $C$ | $f_3(B,C)$ |
|---|---|---|
| $b$ | $c$ | 2 |
| $b$ | $\neg c$ | 4 |
| $\neg b$ | $c$ | 9 |
| $\neg b$ | $\neg c$ | 5 |

| $A$ | $f_1(A)$ |
|---|---|
| $a$ | 1 |
| $\neg a$ | 2 |

Figure 2: Functional definition tables for NSDP example.

## 3 Non-Serial Dynamic Programming

Non-Serial Dynamic Programming (NSDP) as defined in [BB72] involves performing a global operation, usually maximization or minimization, over a series of functions defined on a common domain of discrete variables. To solve a NSDP instance one combines the functions, according to the combination operator, and then performs the desired operation on the resulting much larger function. This process is very expensive; in fact it requires a space equivalent to the cross-product of the variables in the domain.

Fortunately, we can take advantage of Bellman's principle of optimality to reduce the cost of computing the global operation. Bellman's principle states that once all the functions involving a single variable have been combined, we can reduce the size of the resulting interim function by performing the global operation on the interim function. We then carry forth just the values of the variable being removed that produce the best results relative the global operation for each combination of the remaining variables in the function.

For example, suppose that we have a domain of three variables, $\mathcal{V} = \{A, B, C\}$, each of which can take on two states (e.g. $a$ and $\neg a$) and upon which three functions $f_1 : A \to \mathcal{Z}^+$, $f_2 : A, B \to \mathcal{Z}^+$, and $f_3 : B, C \to \mathcal{Z}^+$ are defined. The functions are defined by the tables in Figure 2, and we will assume that we wish to maximize (global operation) the sum (combination operator) of these functions. In this particular case the functions are called the components and their sum is called the objective function [BB72].

If we start by combining $f_1$ and $f_2$ then we would get a function $f_{1\oplus 2}(A, B)$ defined by the table in Figure 3 which can be reduced to $f_{1\oplus 2}(B)$ also seen in Figure 3.

| A | B | $f_{1\oplus 2}(A,B)$ |
|---|---|---|
| a | b | 4 |
| a | ¬b | 5 |
| ¬a | b | 3 |
| ¬a | ¬b | 7 |

| A | B | $f_{1\oplus 2}(B)$ |
|---|---|---|
| a | b | 4 |
| ¬a | ¬b | 7 |

Figure 3: Result of combining $f_1$ and $f_2$.

So when we combine $f_{1\oplus 2}$ with $f_3$ we get only a table based on two variables, $B$ and $C$, with only a note about which value of $A$ maximized $f_{1\oplus 2}$ carried over. It is easy to see that for a larger example the order of combination becomes very important. That is why the secondary problem of NSDP (2-NSDP), that of computing the combination elimination ordering, becomes so important.

In fact the process of computing a solution to 2-NSDP such that the minimum table size is assured is NP-hard [ACP87], with the following variant being known to be NP-Complete.

**Definition 1 (2-NSDP($d$))** *Does there exist a combination — elimination ordering for a set of $n$ function $F = \{f_1, \ldots, f_n\}$ defined over a domain of discrete variables $V$ onto the positive integers s.t. no interim table, before application of Bellman's Principle of Optimality, is formed whose domain contains more than $d$ variables?*

## 4  Optimal Factoring Problem

The optimal factoring problem takes on the same role as 2-NSDP did for NSDP in that it gives us the minimum combination (multiplication)—elimination(marginalization) ordering for the extraction of a joint marginal on a set of target variables, $T$, from a BN. The machinery of the problem is very simple. We start by building a set of sets $S = \{S_1, \ldots, S_m\}$, henceforth to be called the factoring, s.t. each set, $S_i$, is a subset of the variables, $V$, on which the BN is defined.

These sets correspond to the variables in the conditional probability tables for the BN. For example the BN in Figure 1 yields the set representation:

$$S = \{\{A\}, \{A,B\}, \{A,C\}, \{B,C,D\}, \{C,E\}\}$$

The combination of two sets $S_i$ and $S_j$ into a new set $S_{i\oplus j}$ is defined as:

$$S_{i\oplus j} = (S_i \cup S_j) \cap \left( \bigcup_{S_k \in S-\{S_i,S_j\}} S_k \cup T \right)$$

with the cost of the combination $\mu_{S_{i\oplus j}}$ being:

$$\mu(S_{i\oplus j}) = \mu(S_i) + \mu(S_j) + b^{|S_i \cup S_j|}$$

where $b$ is the maximum number of states any single variable in $V$ may take on and $\mu$ is zero for any of the original sets. After creation of the new set $S_{i\oplus j}$ the two original sets, $S_i$ and $S_j$, are removed from $S$ and $S_{i\oplus j}$ is inserted.

In this way the process continues until all the sets have been combined and we are left with just one set equivalent to $T$.

**Definition 2 (Optimal Factoring Problem)** *Given a set $S$ of sets defined over a group of variables $V$ that have no more than $b$ possible states, calculate the combination ordering that for a target set of variables $T$ minimizes the total cost as defined by $\mu$.*

Given a solution to the OFP we can clearly solve a decision problem version:

**Definition 3 (OFP($c$))** *Given a factoring, $S$, defined over a group of variables $V$, a value $b$ to serve as the base of the cost function $\mu$, a target set of variables $T$, and a total cost $c$, does their exist a combination ordering s.t. the cost of deriving $T$ is less than $c$?*

**Theorem 1 (NP-completeness of OFP($c$))** *OFP($c$) is NP-complete.*

Since a solution to the general OFP allows the immediate solution of the decision problem OFP($c$), proof that OFP($c$) is NP-complete shows that the general optimal factoring problem is NP-hard.

## 5  Reduction

We reduce 2-NSDP($d$) to OFP($c$) in the following way:

- Define the variables for OFP($c$) as the variables for 2-NSDP($d$).
- For each function $f_i \in F(1 \leq i \leq n)$ create one set $S_i \in S$ s.t. every variable in the domain of $f_i$ is in the set $S_i$.
- Set $b$, the base of $\mu$, to $n$.
- Set $T = \phi$.
- Set $c = b^{d+1}$

## 6  Proof of Theorem 1

**Definition 4 (Function — Set Correspondence)** *We say that a function $f_i$ corresponds to a set $S_i$ iff the variables in the domain of $f_i$ are equivalent to the members of the set $S_i$.*

**Definition 5 (Function Set – Factoring Equivalence)** *We say that a function set F is equivalent to a factoring S iff for all $S_i \in S$ there exists one and only one corresponding function $f_i \in F$ and there are no unmatched functions in F.*

**Lemma 1 (Combination Set–Function Equivalence)** *Let function set F be equivalent to factoring S. If we combine two sets $S_i \oplus S_j$ in factoring S to get the new factoring $S'$ while combining their corresponding functions $f_i \oplus f_j$ in F to get a new function set $F'$ then $F'$ and $S'$ are function set – factoring equivalent.*

In order to prove Lemma 1 we simply observe that all the sets in $S$ are in a one to one correspondence with domains of all the functions in $F$. Then if we combine any two sets in $S$ and combine their corresponding functions in $F$, before elimination, they are defined on the same variables. That is $S_i \cup S_j$ is defined on the same variables as the domain of $f_i \oplus f_j$.

Furthermore, a variable will be eliminated from $S_i \oplus S_j$ if and only if it is also eliminated from $f_i \oplus f_j$. Since Bellman's principle of optimality only allows variable elimination if the variable exists only in $f_i \oplus f_j$, the combination of $f_i$ with $f_j$ removes the same variables as the removal portion of the set combination rule for the factoring when $T$ is empty.

□

**Lemma 2 (Number of Combinations)** *In either representation there will be exactly $n - 1$ combinations in the solution of the problem.*

Clearly since each combination replaces two sets (functions) with just one there can be no more than $n - 1$ combinations, where $n$ is the number of sets (functions), until there is only one set (function) left.

□

**Lemma 3 (Elimination Equivalence)** *For any sequence of factorings $S^0$, $S^1$, ..., $S^{n-1}$ formed during the solution of the OFP and their equivalent, with respect to which sets (functions) are combined, sequence of function sets $F^0$, $F^1$, ..., $F^{n-1}$ formed during a solution to 2-NSDP, the size of the interim tables formed at each combination is equivalent to the exponent in the cost function of OFP for that combination.*

Note that at the start of the problem we have function set – factoring equivalence, and only corresponding functions and sets are combined in the transition from $F^i$ to $F^{i+1}$ and $S^i$ to $S^{i+1}$ for $0 \leq i \leq n - 2$. Then, by combination function set – factoring equivalence $F^{i+1}$ is function set – factoring equivalent to $S^{i+1}$ after reduction. Furthermore, the number of variables in the set $S_i \cup S_j$ is equivalent to the number of variables in the domain of $f_i \oplus f_j$, before the respective reductions. Thus the dimension of the interim functions is equivalent to the exponent of the cost function.

□

**Lemma 4 (Exhaustive Combination Ordering Equivalence)** *The possible combination orderings for solving the OFP are in one to one correspondence with the possible combination orderings for solving 2-NSDP.*

This follows from the observation that all possible combination sequences for the set of functions have an equivalent factoring elimination ordering and since all elimination sequences for factorings have an equivalent set of function elimination ordering.

□

From Lemma 4 the dimensions of the resulting functions are equivalent to the exponents of the cost functions for any given elimination ordering. Now, note that, if an ordering of sets exists such that the exponent of the cost function $d_1, \ldots, d_{n-1}$ is always below $d$, then the corresponding OFP cost is:

$$\sum_{i=1}^{n-1} b^{d_i} \leq (n-1) * n^d \leq n^{d+1} - n^d < n^{d+1}$$

In other words OFP($c$) answers yes only if there exists a solution for 2-NSDP($d$).

Conversely if every possible combination ordering for 2-NSDP($d$) involves at least one interim table with a dimension of at least $d+1$, then by exhaustive combination ordering equivalence every possible cost for OFP($c$) must exceed $n^{d+1}$ (i.e. at least one term in the summation is greater than or equal to $n^{d+1}$). Thus if there exists no yes solution for 2-NSDP($d$) then there can exist no yes solution for OFP($c$).

This concludes the proof of the reduction portion of Theorem 1. All that remains to establish is that the problem is NP-complete is to show that it is in NP. This is an obvious result since we can check to see if a solution requires fewer that $c$ multiplications in non-deterministic linear time.

We note that the base of the cost function can be reduced to an arbitrary integer $k \geq 2$ by simply replacing each variable in the set of sets with $\lceil \log_k n \rceil$ copies of itself (i.e. $A$ becomes $A_1, \ldots, A_{\lceil \log_k n \rceil}$). Since all these variables will exist in the same sets, they will be eliminated at the same time as the variable in the original set representation would be. Thus we can view the cost at any time for a combination as $k^{\lceil \log_k n \rceil * d_i}$ which is the same as $(k^{\lceil \log_k n \rceil})^{d_i}$ that for the sake of the above proof is equivalent to $n^{d_i}$. It follows, by setting $k = 2$, that $OFP(c)$ is NP-complete even when restricted to instances for which every variable has only two possible values (i.e. $b = 2$).

## 7   Factor Trees

Building upon SPI [LD94] we now present a two stage method for deriving not only the desired joint but also all the single beliefs. The first stage corresponds to the Optimal Factor calculation phase of the Li and D'Ambrosio algorithm and results in the creation of a calculation structure called a factor tree. The second phase involves running a two-stage message passing algorithm on the factor tree to retrieve not only the joint but also all the single beliefs.

The following algorithm constructs a factor tree in four phases.

1. Start by calculating the optimal factoring order for the network given the target set of variables whose joint is desired.

2. From this ordering construct a binary tree showing the combination ordering of the initial probability tables and the conformal tables. A conformal table is defined as any table formed by the combination of two probability tables or the combination of a conformal table with a probability table.

3. Label edges between tables along which variables are marginalized with the variables(s) marginalized before the combination.

4. Add an additional head that has an empty label above the current root, a conformal table labeled with the target set of variables, that has no variables. The edge between the old root and the new is then labeled with the variables in the old root.

Utilizing the above algorithm on the graph shown in Figure 1 factored according to the order seen in equation 4, a factor tree is built that looks like the one in Figure 4.

## 8   Propagation Phase

Once the labeled factor tree described in section 2 is constructed, the algorithm takes on a propagation framework similar to Pearl's method [Pea88] for singly connected networks. We begin at the leaf nodes and propagate up the edges along the direction marked. Messages are tables that are combined using pointwise multiplication [Jen96, Section 4.1].

Once the top of the factor tree is reached we send a new message down the edges in the reverse direction. For the sake of notational similarity we will call the messages that travel up the graph $\lambda$ messages and those that travel down the graph $\pi$ messages. This similarity in naming does not strictly correspond to a similarity in purpose, as we shall soon see.

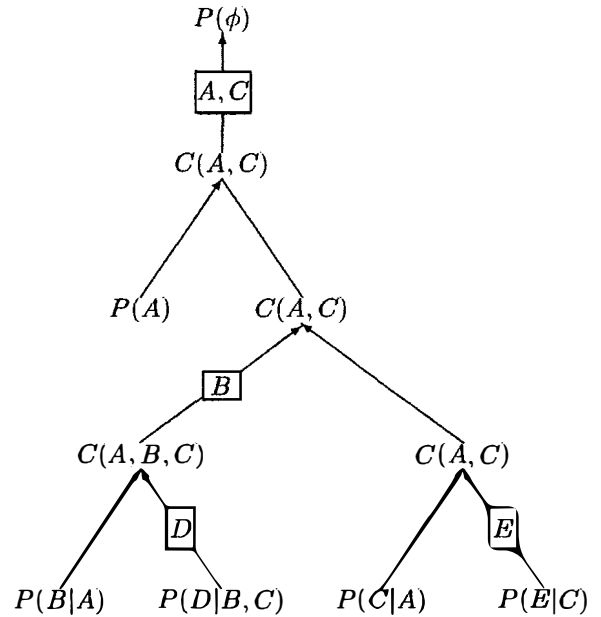

Figure 4: The factor tree for the simple example network.

The following are the procedures performed by each node when it receives a message (either $\lambda$ or $\pi$).

**Leaf Nodes**

$\lambda$ messages — are not received by the leaf nodes by definition.

$\pi$ messages — are ignored by the leaf nodes.

**Root Node**

$\lambda$ message — Set the $\pi$ message for this node to 1 and send it to its child.

**Internal Nodes**

$\lambda$ messages —

1. Store each $\lambda$ message as it arrives.
2. Once both $\lambda$ messages have arrived combine them to create the conformal table for this node.
3. Send the conformal table to the parent as this node's $\lambda$ message.

$\pi$ messages —

1. marginalize away any variables not in the table stored at this node.
2. Combine the $\pi$ message with the $\lambda$ message sent by the left child.
3. Send that as the $\pi$ message to the right child.

 4. Combine the $\pi$ message with the $\lambda$ message sent by the right child.
 5. Send that as the $\pi$ messages to the left child.

The following is the procedure performed by a labeled edge whenever a message is sent along it.

**Labeled Edge**

$\lambda$ message – Store the lambda message in the edge.

$\pi$ message – Combine the $\pi$ message with the stored $\lambda$ message; then marginalize the result onto the variable for which the edge is labeled, obtaining the probability distribution for that variable. In the case of the edge entering into the root it will contain the desired joint.

In the case where variables have been instantiated, marginalization simply passes through the values from the interim table that correspond to the instantiation. In this case $P(\phi)$ will be zero whenever an impossible combination of instantiated variables is given, otherwise it will be the joint marginal probability of the instantiated variables, which is customarily called the probability of the evidence [Jen96, Section 4.2].

## 9 Correctness

Without loss of generality we will prove that the belief in one variable $v$ contained at the edge labeled with $v$ is valid. This edge connects $v_i$ to $v_j$ and we start our proof by removing it from the graph. We then add a new node labeled $v'$ in its place. Two new edges are then added: one from $v_i$ to $v'$ and the other from $v_j$ to $v'$. We then re-orient all other edges in the graph so that $v'$ becomes the root of a new factor tree. Above this node we place a new $P(v)$ node, and we add an edge from $v'$ to the new node $P(v)$ labeled with all the variables contained in $v'$ except $v$. Clearly this is a legal factor tree and represents a legal combination ordering with respect to equations 1 and 2 with respect to distribution.

For example consider the task of retrieving $P(B)$ from the factor tree in Figure 4. Using the above method we modify the tree so that we arrive at the tree shown in Figure 5 which does indeed correspond to the following legal combination ordering

$$P(B) = \sum_{A,C} [[P(B|A) * \sum_D P(D|B,C)] * \\ [[P(\phi) * P(A)] * [P(C|A) * \sum_E P(E|C)]]. \quad (5)$$

In general, consider any labeled edge in the original graph, $G$, and apply a similar transformation to it, obtaining a new graph $G'$. Clearly, the $\pi$ message sent down the edge in the original graph, $G$, is equivalent to one of the $\lambda$ message sent to the node $n'$ in the new graph, $G'$, while the other $\lambda$ message received by $v'$ is the same in both graphs. Thus the edge labeled with $v$ has access to the same messages in the original graph that the node $v'$ has access to in the new graph. Therefore the labeled edge in $G$ can compute the same legal belief in $v$ that $G'$ calculates in the node $v'$. In other words the two messages combined in the labeled edge in the original graph are in actuality the two $\lambda$ messages it would receive in the modified graph, and the belief calculated at the labeled edge is the same as that computed by a factor tree built for the variable in the label.

## 10 Time Complexity

**Define:**

$n$ – the number of variables in the network.

$b$ – the number of states of the largest variable in the network.

$k$ – the number of variables in the largest table in the factor tree.

**multiplications:**

1. Each internal node (of which there are $n-1$) combines 3 tables using no more than $b^k$ multiplications.
   (a) Combines left and right child's $\lambda$ messages into its $\lambda$ message.
   (b) Combines its left child's $\lambda$ message with its parent's $\pi$ message.
   (c) Combines its right child's $\lambda$ message with its parent's $\pi$ message.
2. Each labeled edge (of which there are $n$) combines a $\lambda$ message with a $\pi$ message using no more that $b^k$ multiplications.

**additions:**

1. Each labeled edge marginalizes twice.
   (a) Once whenever a $\lambda$ message passes it using no more than $b^k$ additions.
   (b) Once to remove the final distribution from its stored combination of $\lambda$ and $\pi$ messages using no more than $b^k$ additions.
2. Each internal table where a $\pi$ message is received may need to marginalize the message onto its local label using no more than $b^k$ additions.

This means that the factor tree method to recover the probability of all the variables requires at most $4nb^k$ multiplications and at most $3nb^k$ additions giving the algorithm a total complexity of at most $7nb^k$ significant operations. This time complexity is comparable with the complexity of the tree of cliques approach which runs in at most $5mb^l$ where $m$ is the number of cliques, $b$ is the same, and $l$ is the number of variables in the largest clique [Nea90]. In fact, since the merging of variables into cliques reduces a linear factor, $n$, at the possible expense of an exponential factor, $b^k$, it seems likely that graphs exist for which this algorithm is more efficient (although none have yet been found).

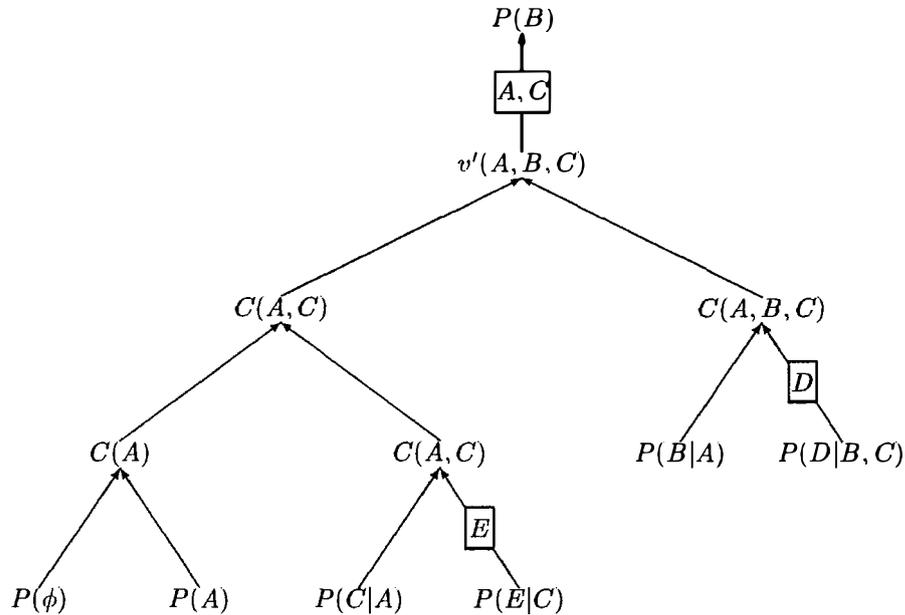

Figure 5: The factor tree for extracting $P(B)$.

Two further improvements can be made to this approach. First, in the case where one wishes to calculate the joint and single beliefs multiple times, one can merge all sub-trees that don't contain a labeled edges into a single node. This merged node then takes the place of the conformal node that was at the root of the merged sub-tree in the factor tree and will save one the amount of calculation that was necessary to build the conformal node that the merged node replaces.

Second, it is interesting to note that the optimal factoring problem can be run with no target set of variables. In this case the set reached just before finishing, or the node just below the root of the factor tree, will contain the most efficiently calculable probability, joint or single, for the network. This fact can be easily established by contradiction: since summing away has no cost for OFP, the joint or marginal immediately beneath the root must be the most efficiently calculable or else a new factoring yielding the more efficiently calculable distribution would yield an OFP solution of less cost.

This would violate the definition of OFP and therefore it is certain that the solution with zero factor set is the most efficient calculation possible. This leads to the interesting observation that the algorithm is calculating the beliefs in all of the single variables in multiplicative constant (approximately 4) time with respect to the most efficient calculation possible for a given network.

## 11 Conclusions

We proved that OFP is NP-hard, confirming Li and D'Ambrosio's [LD94] conjecture. We extended SPI to compute all single-variable marginal beliefs as well as an arbitrary joint belief. The new algorithm contains one NP-hard step, namely the solution of an instance of OFP, thereby reinforcing Jensen and Jensen's [JJ94] conjecture that any scheme for belief updating has an NP-hard optimality step or is less efficient than the junction tree scheme. Three situations are possible:

1. In some cases, the junction tree method is more efficient than the factor tree method described in this paper, and in some cases the factor tree method is more efficient;

2. one method strictly dominates the other;

3. the two methods are of the same complexity.

Determining which of the three conditions hold is a problem for further work. It seems clear, however, that the new algorithm allows for more efficient use of Bayesian networks in systems that require both the joint probability table for some set of variables as well as for all single variables. This is true, because the new approach saves an NP-hard step over using an algorithm from both classes (junction tree and SPI) simultaneously, which is the only other way to derive a target joint as well as the belief in all the single variables without an unpredictable increase in computational cost as required by the variable propagation approach [Jen96, p. 99].

## 12 Acknowledgments

This work was partially supported by the Office of Naval Research project "Dynamic Decision Support for Command, Control, and Communication in the Context of Tactical Defense" (BA97-006). An earlier version of the factor tree method appeared in [Blo98]. Thanks also to Bruce D'Ambrosio for his review and suggestions on an early draft of this paper.